\documentclass{article}

\PassOptionsToPackage{numbers,sort&compress}{natbib}

 \usepackage[preprint]{neurips_2025}

\usepackage[utf8]{inputenc} 
\usepackage[T1]{fontenc}    
\usepackage{hyperref}       
\usepackage{url}            
\usepackage{booktabs}       
\usepackage{amsfonts}       
\usepackage{nicefrac}       
\usepackage{microtype}      
\usepackage{xcolor}         
\usepackage{amsmath}
\usepackage{array}
\usepackage{tabularx}
\usepackage[table]{xcolor}
\usepackage{makecell}
\usepackage{graphicx}
\usepackage{float}

\newcolumntype{L}{>{\raggedright\arraybackslash}p{0.12\textwidth}}
\newcolumntype{M}{>{\centering\arraybackslash\scriptsize}X}

\definecolor{headergray}{RGB}{245,245,245}

\title{Beyond Next-Token Prediction: 
\\
A Performance Characterization of 
\\
Diffusion versus Autoregressive Language Models
}

\author{
Minseo Kim\textsuperscript{1}\quad
Coleman Hooper\textsuperscript{2}\quad
Aditya Tomar\textsuperscript{2}\quad
Chenfeng Xu\textsuperscript{2,5}\\
\textbf{Mehrdad Farajtabar\textsuperscript{6}\quad
Michael W. Mahoney\textsuperscript{2,3,4}\quad
Kurt Keutzer\textsuperscript{2}\quad
Amir Gholami\textsuperscript{2,3}}\\[0.35em]
\textsuperscript{1}Seoul National University\quad
\textsuperscript{2}University of California, Berkeley\quad
\textsuperscript{3}ICSI\quad
\textsuperscript{4}LBNL \\
\textsuperscript{5}University of Texas at Austin\quad
\textsuperscript{6}Apple\thanks{advisory role }
}

\usepackage[textwidth=2.5cm]{todonotes}

\begin{document}

\maketitle

\vspace{-2mm}
\begin{abstract}
Large Language Models (LLMs) have achieved state-of-the-art performance on a broad range of Natural Language Processing (NLP) tasks, including document processing and code generation.
Autoregressive Language Models (ARMs), which generate tokens sequentially conditioned on all previous tokens, have been the predominant paradigm for LLMs.
While these models have achieved high accuracy across a range of downstream tasks, they exhibit low arithmetic intensity due to the inherent sequential dependency in next-token prediction.
Recently, Diffusion Language Models (DLMs) have emerged as a promising alternative architecture.
DLMs generate output tokens \textit{in parallel}, mitigating the limitations of sequential decoding.
However, the performance implications of DLMs relative to commonly deployed ARMs are not fully understood.
In this work, we present a comprehensive study of the performance characteristics of ARMs and DLMs, combining theoretical analysis with empirical profiling to characterize the trade-offs between these approaches.
We show that although DLMs can achieve higher arithmetic intensity than ARMs by leveraging parallelism across token positions, they fail to scale effectively with longer contexts.
We then explore block-wise decoding for DLMs, which decouples arithmetic intensity from sequence length and enables better scaling to long contexts (similar to ARMs).
We also examine batched inference and find that ARMs exhibit superior throughput as they benefit more from parallelism across sequences in the batch.
Finally, we highlight opportunities for accelerating DLM inference, emphasizing that reducing the number of sampling steps is key for open-source DLMs to achieve lower latency relative to ARMs.
\end{abstract}

\section{Introduction}
\label{sec:introduction}

Large Language Models (LLMs) have ushered in a new era for machine learning and artificial intelligence, demonstrating broad applicability across instruction following, code generation, summarization, and multi-turn dialogue~\cite{openai2023gpt4}. 
Modern LLMs are predominantly Autoregressive Language Models (ARMs): decoder-only Transformers trained with a next-token prediction objective under a causal attention mask~\cite{zhang2022opt, dubey2024llama}. 
Although the training process is highly parallelizable thanks to teacher forcing, this objective imposes a strict sequential dependency at inference time. 
To mitigate this, various techniques have emerged, with speculative decoding being a prominent example that uses a smaller draft model to generate token bursts for parallel verification~\cite{kim2023speculative,chen2023accelerating,leviathan2023fast}. However, the drafting path remains sequential and offers limited speedups.

As a promising alternative, Diffusion Language Models (DLMs) have recently gained attention for their new possibilities in text generation. Originating from the broader success of diffusion models in modalities like image, audio, and video synthesis~\cite{ho2020ddpm,ramesh2022hierarchical,chen2021wavegrad,esser2023sceneguided}, DLMs operate by iteratively denoising a sequence of random noise into coherent text~\cite{li2022diffusionlm, gong2023diffuseq, strudel2022sed}. 
This process updates all token positions \textit{in parallel} at each step, thereby breaking the token-level sequential dependency inherent to autoregressive decoding, and showing promise as an architecture that better exploits parallelism along the sequence dimension. 
Closed-source DLMs have publicly emphasized substantial speedups (e.g., Google’s experimental Gemini Diffusion~\cite{deepmind2025geminidiffusion} and Inception Labs’ Mercury~\cite{khanna2025mercury} claim up to \emph{10$\times$} faster and cheaper generation than strong autoregressive baselines), whereas current open-source DLMs often remain substantially slower than comparable ARMs in practice (see Section~\ref{subsec:runtime-analysis}).

In this work, we present a systematic performance characterization of diffusion-based text generation, relative to standard autoregression, to identify directions for closing the gap with closed-source DLMs. 
Our contributions are threefold:

\begin{itemize}
    \item 
    \textbf{Unified analysis.} 
    (Sections \ref{subsec:asymptotic_analysis_of_arithmetic_intensity} and \ref{subsec:runtime-analysis}) 
    We present profiling data along with roofline modeling analysis to characterize ARM prefill/decode and DLM variants across prompt and output lengths.
    Our analysis shows that DLMs have greater arithmetic intensity (AI) than ARMs during decoding, but scale poorly to long contexts due to their need to re-process the full sequence at each sampling step.
    \item 
    \textbf{Block-wise decoding and batching.} 
    (Sections \ref{subsec:blockwise} and \ref{subsec:batched-inference}) 
    We demonstrate how block-wise DLMs reduce AI relative to naive DLM decoding, while maintaining higher AI than ARMs and scaling well to long context lengths.
    In contrast, for batched serving, ARMs exhibit superior throughput, as they better exploit sequence-level parallelism compared to DLMs (with or without block-wise decoding).
    \item 
    \textbf{Directions to close the gap.} 
    (Section \ref{sec:discussion}) 
    We surface concrete strategies to narrow the current gap with closed-source DLMs, highlighting how reducing the number of diffusion steps is critical for improving the latency of open-source DLMs relative to open-source ARMs.
\end{itemize}
\section{Background}
\label{sec:background}

Here, we describe several lines of related background work.

\textbf{Autoregressive and Diffusion Language Models.}
Autoregressive language models (ARMs), the foundation of modern LLMs, are trained with a next-token prediction objective under a causal mask, restricting attention to past tokens~\cite{radford2019language,brown2020language}. 
While training can be efficiently parallelized, inference is inherently sequential because each new token depends on the entire previously generated prefix~\cite{vaswani2017attention}. 
This sequential dependency creates a performance bottleneck, leading to low arithmetic intensity and a decode path that is memory-bandwidth-bound~\cite{kim2023squeezellm,kim2023full}. 
To avoid recomputing activations for past tokens at every step, \emph{KV caching} has become standard. 
By storing Key (K) and Value (V) vectors during decoding, it largely eliminates recomputation and accelerates autoregressive inference~\cite{pope2022efficient}. 
The trade-off is a substantial increase in memory footprint and bandwidth pressure~\cite{hooper2024kvquant}.

Diffusion models generate samples by initializing from Gaussian noise and iteratively applying reverse sampling steps, gradually transforming noise into a clean sample from the data distribution~\cite{ho2020ddpm}. 
This family of models has been successfully adopted across modalities, including text-conditioned image generation~\cite{ramesh2022hierarchical}, audio synthesis~\cite{chen2021wavegrad}, video generation~\cite{esser2023sceneguided}, and even language.
In language modeling, diffusion language models (DLMs) generate over a fixed-length sequence via iterative refinement, updating all token positions in parallel at each step~\cite{li2022diffusionlm,gong2023diffuseq,strudel2022sed}. Architecturally, DLMs typically adopt Transformer architectures with \textit{bidirectional} (non-causal) attention, allowing each position to attend to the full context~\cite{nie2025large}.
Early work focused on continuous DLMs that operate in embedding space~\cite{gao2024embedding,gong2023diffuseq,strudel2022sed}; more recently, masked diffusion models (MDMs), a class of discrete DLMs that operate directly on categorical token spaces, have emerged as competitive alternatives to ARMs~\cite{dream2025,nie2025large}. A key systems challenge for DLMs is that bidirectional attention is incompatible with standard KV caching, causing latency to increase substantially as context length grows (see Section~\ref{subsec:runtime-analysis}). 
To mitigate this, \textit{block-wise decoding} divides the sequence into blocks. Decoding proceeds in an autoregressive manner across blocks, while positions within each block are updated in parallel, enabling approximate KV caching of the prompt and non-active blocks.~\cite{wu2025fastdllm,ma2025dkvcache,hu2025freecache}.
A separate line of work, speculative diffusion decoding, keeps the main model autoregressive while employing a diffusion model as the drafter~\cite{christopher2024specdiff}.

\textbf{Arithmetic Intensity and Roofline Model.}
While many analyses focus solely on the raw FLOP count of an algorithm, this view can be misleading. 
Kernel runtime may instead be dictated by the number of memory operations required if the kernel is bottlenecked by memory traffic.
Thus, considering arithmetic intensity (AI) relative to memory bandwidth is essential for understanding true system performance.
AI quantifies how much computation is performed per byte moved; namely, the number of floating-point operations (FLOPs) per byte (or, approximately, per memory operation, MOP):
\begin{equation*}
\mathrm{Arithmetic~Intensity~(AI)} \;=\; \frac{\#\mathrm{FLOPs}}{\#\mathrm{Bytes~moved}} \;\approx\; \frac{\#\mathrm{FLOPs}}{\#\mathrm{MOPs}}
\end{equation*}
Low AI indicates memory-bound behavior (limited by peak memory bandwidth), whereas high AI indicates compute-bound behavior (limited by peak FLOP/s).

One method to visualize bottlenecks across prompt lengths, KV caching, and batch sizes is the roofline model~\cite{williams2009roofline}.
The roofline model provides a visual criterion for classifying programs as compute-bound or memory-bound via the hardware ceilings for compute and bandwidth. 
The ridge point is the knee where the bandwidth roof meets the compute roof, occurring at \(\mathrm{AI}_{\text{ridge}} = \frac{\text{peak compute}}{\text{peak memory bandwidth}}\) (FLOP/byte); workloads with \(\mathrm{AI}<\mathrm{AI}_{\text{ridge}}\) are memory-bound, and those with \(\mathrm{AI}\ge\mathrm{AI}_{\text{ridge}}\) are compute-bound.

\vspace{-1mm}
\section{Workload Analysis}
\label{sec:analysis}

\vspace{-1mm}
In this section, we present our main results.

\vspace{-1mm}
\subsection{Experimental Setup}
\vspace{-1mm}

\textbf{Models.}
We study the LLaMA-3-8B-Instruct model~\cite{dubey2024llama} (an ARM) and the LLaDA-8B-Instruct model~\cite{nie2025large} (a DLM) as representative baselines
(as they have comparable model size and comparable downstream task performance as reported in \cite{nie2025large}). 
For block-wise DLM experiments with approximate KV caching (outlined in Section \ref{subsec:blockwise}), we adopt the LLaDA variant (dual cache) from Fast-dLLM~\cite{wu2025fastdllm}, as the original LLaDA does not support KV caching.

\textbf{Methodology.}
We extend prior analytical modeling frameworks~\cite{kim2023full,tiwari2025quantspec} to (i) decompose ARM inference into prefill and decode, and (ii) model DLM inference as multi-step refinement with optional block-wise KV reuse.

\textbf{Hardware.} 
We perform measurements using a single NVIDIA RTX~A6000 (48GB) with FP16 inference under PyTorch/CUDA (unless otherwise stated).
Roofline parameters follow the vendor specifications.

\subsection{Asymptotic Analysis of Arithmetic Intensity in ARM and DLM}
\label{subsec:asymptotic_analysis_of_arithmetic_intensity}
\vspace{-1mm}

\begin{table}[t]
  \centering
  \caption{Asymptotic FLOPs, memory operations (MOPs), and arithmetic intensity (AI) for ARM prefill/decode and DLM variants. Here $L=L_p+L_g$ (prompt + generated length), $B$ is batch size, $d$ is hidden dimension, $K$ is the number of diffusion steps, $G$ is the block size, and $N=\lceil L_g/G\rceil$ is the number of blocks. The first and second terms in each entry correspond to linear/attention operations for FLOPs, and weight/activation loads for MOPs, respectively.
  }
  \label{tab:ai_arlm_dlm}

  \setlength{\tabcolsep}{4pt}
  \renewcommand{\arraystretch}{1.22}

  \begin{tabularx}{\textwidth}{L | M | M | M | M}
    \toprule
    \rowcolor{headergray}
    & \makecell[c]{\textbf{ARM}\\\textbf{Prefill}}
    & \makecell[c]{\textbf{ARM}\\\textbf{Decode}}
    & \makecell[c]{\textbf{Block-wise DLM}\\\textbf{(KV Cache)}}
    & \makecell[c]{\textbf{Naive DLM}\\\textbf{(no KV Cache)}} \\
    \midrule

    \textbf{FLOPs} &
    $\begin{aligned}[t]
      \mathcal{O}(B L_p d^{2}) \\
      +\ \mathcal{O}(B L_p^{2} d)
    \end{aligned}$ &
    $\begin{aligned}[t]
      \mathcal{O}(B L_g d^{2}) \\
      +\ \mathcal{O}(B L_g L_p d)
    \end{aligned}$ &
    $\begin{aligned}[t]
      \mathcal{O}(K B G d^{2}) \\
      + \mathcal{O}(K B G L d)
    \end{aligned}$ &
    $\begin{aligned}[t]
      \mathcal{O}(K B L d^{2}) \\
      +\ \mathcal{O}(K B L^{2} d)
    \end{aligned}$ \\

    \rowcolor{gray!40}
    \textbf{MOPs} &
    $\begin{aligned}[t]
    \mathcal{O}(d^{2}) \\
      +\ \mathcal{O}(B L_p d)
    \end{aligned}$ &
    $\begin{aligned}[t]
      \mathcal{O}(L_g d^{2}) \\
      +\ \mathcal{O}(B L_g L_p d)
    \end{aligned}$ &
    $\begin{aligned}[t]
      \mathcal{O}(K d^{2}) \\
      +\ \mathcal{O}(K B L d)
    \end{aligned}$ &
    $\begin{aligned}[t]
      \mathcal{O}(K d^{2}) \\
      +\ \mathcal{O}(K B L d)
    \end{aligned}$ \\

    \textbf{AI} &
    $\begin{aligned}[t]
      \approx \begin{cases}
                \mathcal{O}(B L_p) , ~ L_p \ll d \\
                \mathcal{O}(L_p),~~~~ L_p \gg d
            \end{cases}
    \end{aligned}$ &
    $\begin{aligned}[t]
      \approx \begin{cases}
                \mathcal{O}(B), & L_p \ll d \\
                \mathcal{O}(1), & L_p \gg d
            \end{cases}
    \end{aligned}$ &
    $\begin{aligned}[t]
        \approx \begin{cases}
                \mathcal{O}(B G), & L_p \ll d \\
                \mathcal{O}(G), & L_p \gg d
            \end{cases}
    \end{aligned}$ &
    $\begin{aligned}[t]
        \approx \begin{cases}
                \mathcal{O}(B L), & L_p \ll d \\
                \mathcal{O}(L), & L_p \gg d
            \end{cases}
    \end{aligned}$ \\
    \bottomrule
  \end{tabularx}
\end{table}

Table~\ref{tab:ai_arlm_dlm} shows asymptotic analysis of arithmetic intensity (AI) for ARMs and DLMs.

\textbf{ARMs.} 
Prefill shifts from \emph{linear-projection}-dominated AI ($\mathcal{O}(B L_p)$) when $L_p \ll d$ to \emph{attention}-dominated AI ($\mathcal{O}(L_p)$) when $L_p \gg d$. 
During the subsequent decode phase, each step involves small matrix multiplications (\emph{linear projections}) and retrieval of the growing KV cache for \emph{attention}. As a result, the AI is initially $\mathcal{O}(B)$ when $L_p \ll d$, but it drops to $\mathcal{O}(1)$ once $L_p \gg d$, since the memory load from attention begins to dominate the fixed per-token compute; see also~\cite{tiwari2025quantspec}.

\textbf{Block-DLMs.} 
In a block-wise DLM with block size $G$, each refinement step updates a block of $G$ tokens in parallel. The parallel computation within its linear layers is analogous to an ARM prefill operation on a sequence of length $G$. However, the attention mechanism is fundamentally distinct, as these G tokens attend to the KV cache representing the entire sequence $L$. The cost of initializing and periodically updating this cache via a full-attention pass is omitted, as it is a less dominant term compared to the cumulative cost of the $K$ refinement steps.

\textbf{Naive DLMs.}
In contrast, a naive (fully bidirectional) DLM processes the full sequence each step; the scale factor is the total length $L=L_p+L_g$, giving $\mathrm{AI} \approx \mathcal{O}(B L)$ for $L_p \ll d$ and $\mathcal{O}(L)$ for $L_p \gg d$. Notably, when KV caching is absent, each sampling step recomputes the full sequence, akin to ARM prefill, and is therefore typically compute-bound.

\vspace{-1mm}
\subsection{Runtime and Roofline Analysis of ARM and DLM}
\label{subsec:runtime-analysis}
\vspace{-1mm}

\begin{figure}[t]
  \centering
  \includegraphics[width=0.96\linewidth]{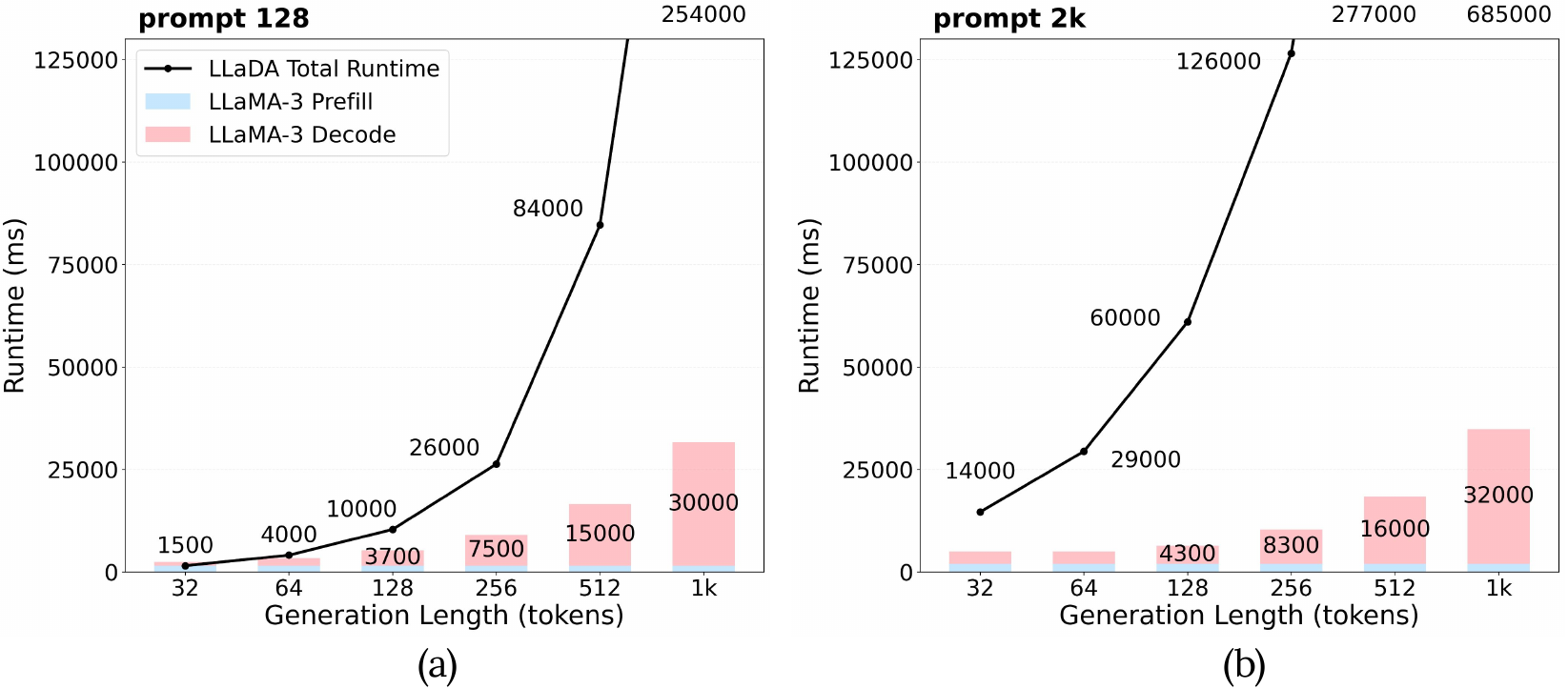}
  \vspace{-2mm}
  \caption{\textbf{Runtime comparison of ARM and DLM.}
  (a) End-to-end runtime vs.\ generation length \(L_g\) for a short context (\(L_p=128\)); bars show ARM prefill/decode and the line shows naive DLM total.
  (b) Same as (a) but for a long context (\(L_p=2\mathrm{k}\)).
  }
  \label{fig:runtime-analysis}
\end{figure}
\begin{figure}[t]
  \centering
  \includegraphics[width=0.5\linewidth]{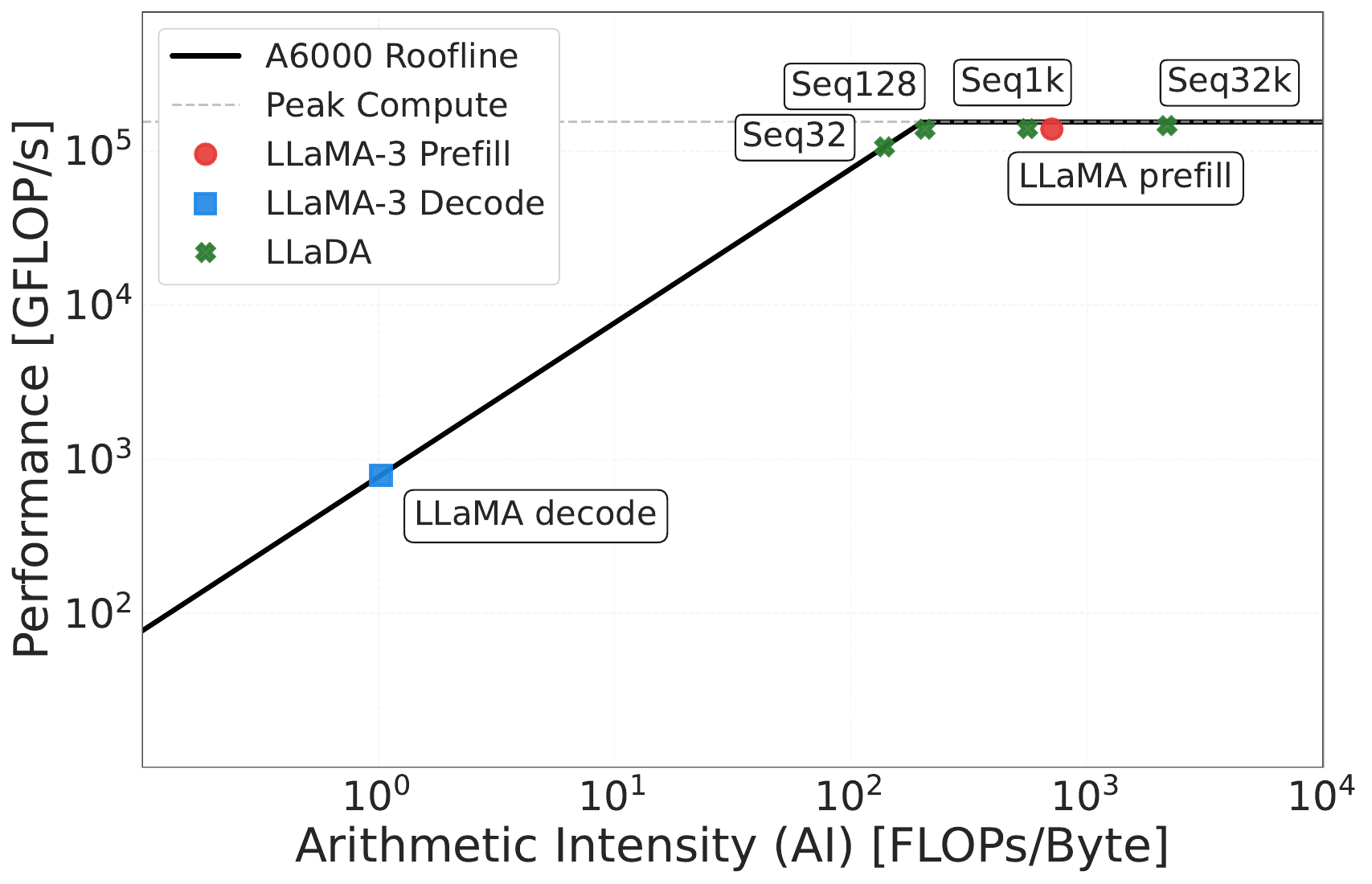}
  \vspace{-1mm}
  \caption{\textbf{Roofline comparison of ARM and DLM.}
  Roofline placement across sequence lengths for ARM prefill, ARM decode, and naive DLM, simulated for an NVIDIA RTX A6000 with a peak performance of 154.8 TFLOP/s and a ridge point at an AI of 201.6 FLOP/Byte.
  }
  \label{fig:roofline-analysis}
\end{figure}

We study end-to-end runtime for ARMs and DLMs as generation length \(L_g\) varies under two prompt lengths, \(L_p{=}128\) (short) and \(L_p{=}2\mathrm{k}\) (long).
We use batch size \(B{=}1\), do \emph{not} use block-wise decoding and approximate caching, and set the number of diffusion steps to match the generation length, \(K {=}L_g\), a common convention in diffusion language models.
The roofline analysis assumes a peak FP16 performance of 154.8 TFLOP/s for the A6000 GPU (see Appendix~\ref{appendix:a6000-roofline} for details) and models LLaMA-3 with a 2k-token sequence length.

Figure~\ref{fig:runtime-analysis}~(a) shows latency for the short prompt. 
The ARM runtime decomposes into prefill (blue), which is flat in \(L_g\), and decode (red), which grows modestly because KV caching prevents recomputation. 
The DLM is quite fast for short prompts and small $L_g$, thanks to parallel token updates, but its latency rises sharply as \(L_g\) grows. 
With a long prompt in Figure~\ref{fig:runtime-analysis}~(b), DLM latency is higher even at small \(L_g\), since each sampling step recomputes the entire sequence, whereas ARM decode reuses the cached tokens from prior steps.

Figure~\ref{fig:roofline-analysis} places these workloads on the roofline. ARM prefill (red) lies to the right of the ridge point and is \textit{compute-bound}, reflecting large GEMMs over the full prompt. ARM decode (blue) sits on the bandwidth roof and is \textit{memory-bound}, dominated by KV-cache traffic. For naive DLM (green), short sequences (e.g., up to 128 tokens) lie near the bandwidth roof, while longer sequences cross the ridge point and become \textit{compute-bound} as full-sequence denoising dominates.

\vspace{-1mm}
\subsection{Implications of Block-wise Decoding with KV Cache}
\label{subsec:blockwise}
\vspace{-1mm}

\begin{figure}[t]
  \centering
  \includegraphics[width=\linewidth]{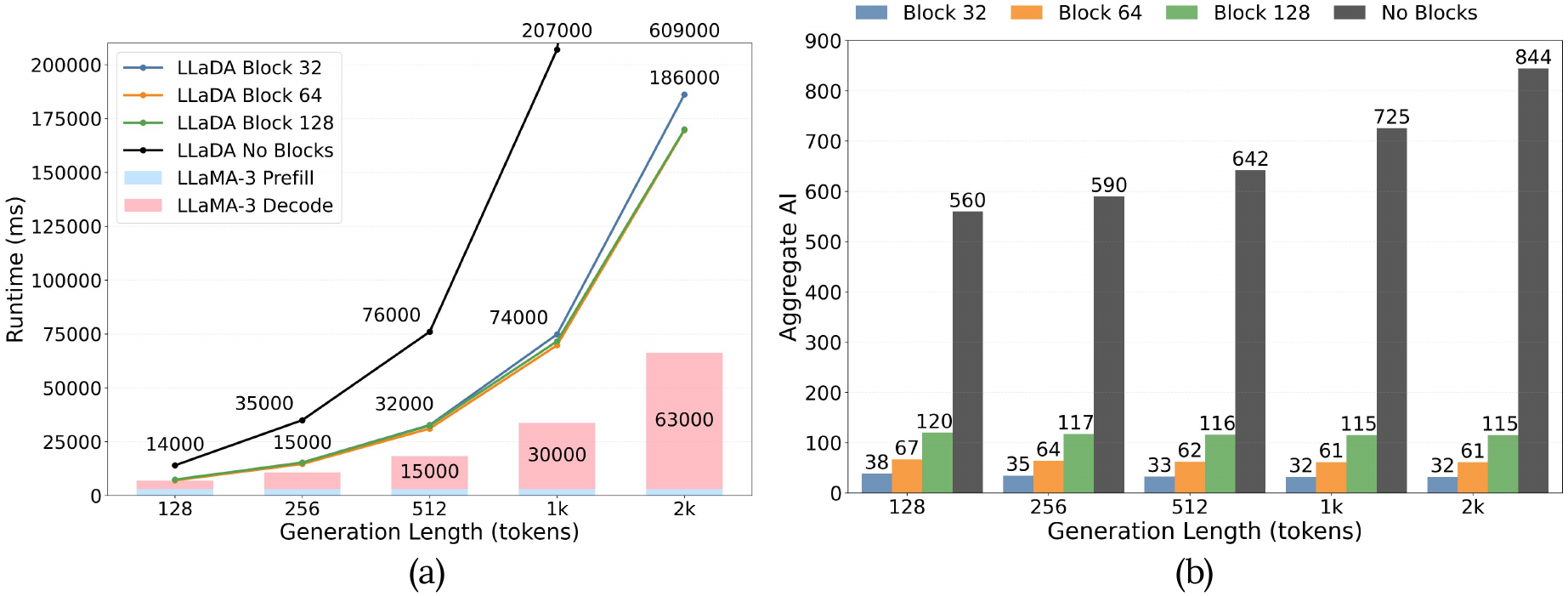}
  \vspace{-6mm}
  \caption{
  \textbf{Runtime and AI for naive and block-wise DLM with KV caching.}
    (a) End-to-end runtime vs. generation length \(L_g\) at a fixed prompt \(L_p=1\mathrm{k}\); bars show ARM (for reference); lines show the naive DLM ("No Blocks") and block-wise DLM for block sizes \(G\in\{32,64,128\}\) with diffusion steps \(K=L_g\).
    (b) Aggregate AI vs. \(L_g\) for the naive and block-wise DLM configurations.
  }
  \label{fig:block-wise-analysis}
\end{figure}

\textit{Block-wise decoding} combines the strengths of ARMs and DLMs: it reuses the KV cache for the prompt and non-active blocks (ARM-like), while updating all tokens within the active block in parallel (DLM-like). 
The diffusion steps $K$ are typically distributed evenly across blocks.
We conduct a profiled analysis, where we fix the prompt length to \(L_p=1\mathrm{k}\), consider block sizes \(G\in\{32,64,128\}\), and set the number of diffusion steps equal to the generation length (\(K=L_g\)).
Figure~\ref{fig:block-wise-analysis}~(a) compares end-to-end latency versus generation length $L_g$, demonstrating how block-wise decoding with KV caching reduces latency by roughly \(2\times\) to \(3\times\), compared to naive DLM (labeled "No Blocks"). 
Moreover, Figure~\ref{fig:block-wise-analysis}~(b) shows that this approach not only reduces the AI but also makes it invariant to the generation length, depending only on the block size \(G\). 
This reduction in AI translates into lower latency in compute-bound settings (e.g., naive DLM inference in Fig.~\ref{fig:block-wise-analysis}~(b)), and block-wise decoding reduces the required FLOPs relative to naive DLM (as highlighted in Table~\ref{tab:ai_arlm_dlm}).
Furthermore, block-wise decoding preserves (and can even improve) task accuracy despite these efficiency gains; we report accuracy results in Appendix~\ref{appendix:blockwise-decoding}.

Note that even with KV caching, the DLM remains slower than the ARM at matched prompt and output lengths when \(K\) scales with \(L_g\). For instance, the widely used configuration of \(K=L_g\)~\cite{nie2025large, dream2025} implies that the model denoises and finalizes only one token per step on average, which is hugely inefficient. This underscores the importance of \textit{reducing diffusion steps} via efficient multi-token denoising, a point we revisit in Section~\ref{sec:discussion}.

\vspace{-1mm}
\subsection{Implications of Inference with Different Batch Sizes}
\label{subsec:batched-inference}
\vspace{-1mm}

\begin{figure}[t]
  \centering
  \includegraphics[width=\linewidth]{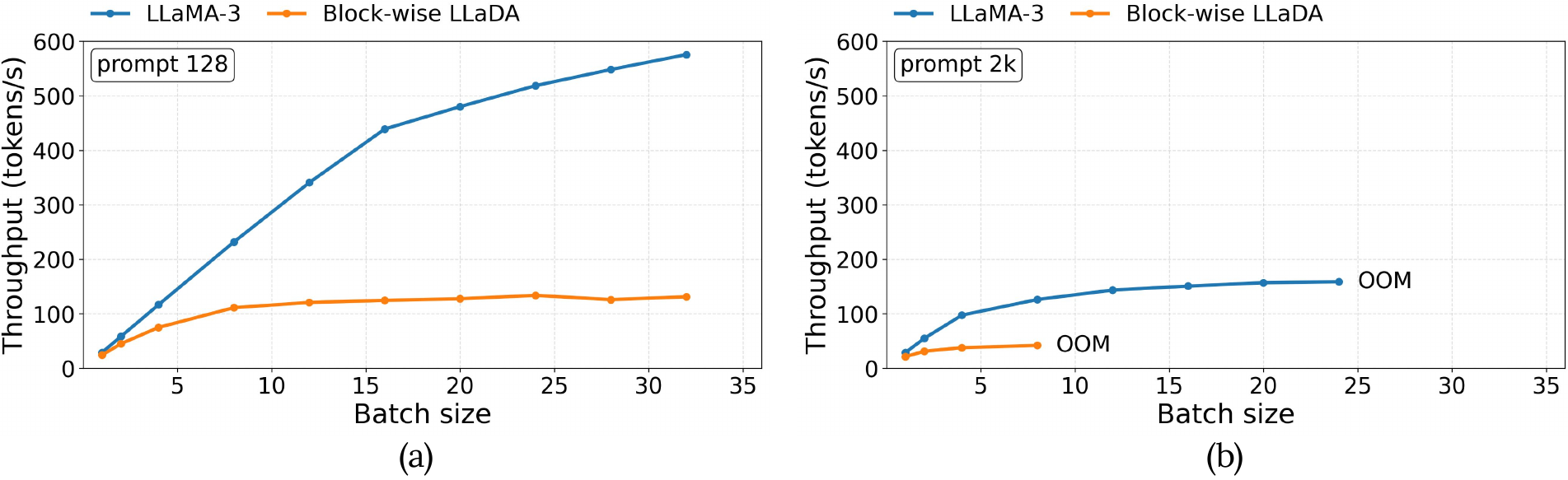}
  \vspace{-6mm}
  \caption{
    \textbf{Throughput vs.\ batch size for ARM and block-wise DLM.}
    Throughput (tokens/s) as a function of batch size $B$ for (a) short prompt ($L_p=128$) and (b) long prompt ($L_p=2$k). The generation length is fixed at $L_g=128$, and OOM denotes out-of-memory.
  }
  \label{fig:batched-runtime-analysis}
\end{figure}

Real-world applications often serve multiple users concurrently, making scalability with respect to the batch size $B$ a critical metric. 
We therefore measure throughput (tokens/s) as a function of $B$, fixing the generation length at $L_g=128$ and evaluating two prompt categories: short ($L_p=128$) and long ($L_p=2\mathrm{k}$). 
We set the block size to $G=32$ and use $K=L_g$ for the block-wise DLM. 
All experiments in this section are run on a single NVIDIA A100 (80 GB) GPU to accommodate the memory demands of batched inference. 
We omit the naive DLM, as prior work shows it does not benefit from batching due to its consistently \textit{compute-bound} bidirectional full attention~\cite{wu2025fastdllm}.

Figure~\ref{fig:batched-runtime-analysis} indicates that for both ARMs and DLMs, throughput scaling with batch size diminishes as prompt length $L_p$ increases.
As $L_p$ grows, the memory-intensive task of reading a large \textit{KV cache} becomes the dominant part of the workload, limiting the potential throughput gains from batching.
Despite this shared trend, ARMs consistently exhibit better batching scalability than DLMs across all prompt lengths.
For example, at $L_p=128$, ARM throughput scales effectively up to a batch size of 16, whereas DLM throughput already plateaus around 8 and remains much lower in absolute terms. 
This disparity arises because the ARM's lightweight, memory-bound decoding effectively leverages batch parallelism, while the DLM's throughput is constrained by two factors: its compute-heavy block generation leads to an earlier performance plateau, and the high cumulative cost of iterative refinement results in a lower overall peak.

The plots also reveal memory limitations. For the block-wise DLM, out-of-memory (OOM) errors appear at smaller $B$ as $L_p$ grows. 
Even though decoding proceeds block by block, attention within each block is bidirectional and requires the KV cache for the full sequence length $L = L_p + L_g$ to be resident. 
Consequently, peak memory scales with $B \times L$, causing longer prompts to hit OOM at smaller batch sizes.

\vspace{-1mm}
\section{Discussion and Future Work}
\label{sec:discussion}
\vspace{-1mm}

Autoregressive language models (ARMs) have driven major progress in NLP, but their next-token objective imposes sequential dependencies that increase latency and can degrade quality by limiting bidirectional context. 
Diffusion language models (DLMs) are a promising alternative, as they have potential to decode many tokens in parallel, while conditioning on the full context. 
We have presented a comprehensive analysis of the performance trade-offs between open-source ARMs and DLMs.

Our analysis shows that a naive DLM exhibits substantial latency overheads relative to ARMs as the context length increases.
To ameliorate this, block-wise decoding with approximate KV caching offers an effective hybrid approach.
This method achieves increased arithmetic intensity (AI) by exploiting intra-block parallelism, while still scaling well to long contexts, similar to ARMs.
However, in batched serving, the block-wise DLM's parallel generation saturates the GPU at smaller batch sizes than ARMs.
Combined with the DLM's high cost of iterative refinement, this results in ARMs providing distinct throughput advantages in the large-batch regime.
Recent closed-source systems~\cite{deepmind2025geminidiffusion,inceptionlabs2025bedrock,khanna2025mercury} report striking latency gains at comparable quality, even faster than ARMs.
Based on our analysis, we therefore conclude that two core bottlenecks are responsible for the slowdown of current open-source DLMs: poor scaling with context length and high refinement costs.

For the scaling problem, Section~\ref{subsec:blockwise} confirms that block-wise decoding with approximate KV caching~\cite{wu2025fastdllm,ma2025dkvcache,hu2025freecache} is an indispensable strategy.
Yet, to truly rival ARM speed, the most critical path forward involves reducing the number of sampling steps $K$.
In common configurations, $K$ scales with the generation length $L_g$, leading to high per-output costs and compute saturation, as highlighted in Section~\ref{subsec:batched-inference}.
Recent approaches attempt to cut steps without quality loss, such as multi-token finalization with autoregressive guidance~\cite{hu2025freecache}, confidence-based early finalization~\cite{wu2025fastdllm}, and distillation into few-step or even one-step student models~\cite{chen2025dlmone}.
Reducing sampling steps is particularly promising in \textit{small-batch} scenarios, where a DLM's throughput could surpass an ARM's, as mentioned in Appendix~\ref{appendix:reduce-steps}.
Finally, orthogonal system-level optimizations can offer additional speedups. These include compute-focused methods like low-precision execution as well as sparse attention methods that utilize only the most salient cache entries.

\subsection*{Acknowledgements}
We acknowledge gracious support from the FuriosaAI, Intel, Apple, NVIDIA, Macronix, and Mozilla team.
We also appreciate the support from Microsoft through their Accelerating Foundation Model Research.
Furthermore, we appreciate support from
Google Cloud, the Google TRC team and Prof. David Patterson.
Prof. Keutzer's lab is sponsored by the Intel corporation, UC Berkeley oneAPI Center of Excellence, Intel VLAB team, as well as funding through BDD and BAIR.
MWM also acknowledges DARPA, NSF, the DOE Competitive Portfolios grant, and the DOE SciGPT grant.
Our conclusions do not necessarily reflect the position or the policy of our sponsors, and no official endorsement should be~inferred.

\bibliographystyle{plainnat} 
\bibliography{references}


\newpage
\appendix

\section{RTX A6000 Roofline Parameters}
\label{appendix:a6000-roofline}

We derive the roofline parameters for the NVIDIA RTX~A6000 (GA102).
We assume \emph{dense} FP16 Tensor Core computation (without 2:4 structured sparsity) and use the GPU boost clock.
Hardware specifications (SM count, Tensor Cores per SM, boost clock, memory bandwidth,
and FP16 per-Tensor Core throughput) are taken from the
GA102 architecture whitepaper.
\footnote{NVIDIA Ampere GA102 Architecture Whitepaper: \url{https://www.nvidia.com/content/PDF/nvidia-ampere-ga-102-gpu-architecture-whitepaper-v2.pdf}.}

\paragraph{Peak FP16 Performance.}
\[
\begin{aligned}
\text{Peak FP16 (dense)}
&= 84~\mathrm{SM}
   \times 4~\mathrm{Tensor~Core/SM}
   \times 128~\mathrm{FMA/(cycle\cdot Tensor~Core)} \\
&\quad \times 1.8~\mathrm{GHz}
   \times 2~\mathrm{FLOP/FMA} \approx 154.8~\mathrm{TFLOP/s}.
\end{aligned}
\]

\paragraph{Ridge Point.}
With memory bandwidth \(768~\mathrm{GB/s}\), the ridge arithmetic intensity is
\[
\mathrm{AI}_{\mathrm{ridge}}
= \frac{154.8~\mathrm{TFLOP/s}}{768~\mathrm{GB/s}}
\approx 201.6~\mathrm{FLOP/Byte}.
\]

\section{Impact of Block-wise Decoding on Generation Quality}
\label{appendix:blockwise-decoding}

\begin{table}[t]
  \centering
  \caption{
    DLM performance on GSM8K and HumanEval with and without block-wise decoding.
    We report results for block sizes $G\in\{64,32,16,8\}$ and the non-block-wise baseline (\texttt{None}).
    The generation length is fixed to $L_g=128$.
    GSM8K is reported as exact-match (EM) accuracy, and HumanEval is reported as pass@1.
  }
  \label{tab:diff-blocks-acc}
  \begin{tabular}{c|cc}
    \toprule
    \rowcolor{headergray}
    Block Size $G$ & GSM8K (4-shot, EM) & HumanEval (0-shot, pass@1) \\
    \midrule
    None & 63.2 & 13.4 \\
    64  & 73.8 & 34.1 \\
    32  & 73.6 & 35.4 \\
    16  & 75.7 & 38.4 \\
    8   & 75.0 & 38.4 \\
    \bottomrule
  \end{tabular}
\end{table}

As discussed in Section~\ref{subsec:blockwise}, block-wise decoding combined with KV caching provides systematic efficiency benefits by reducing redundant computation, thereby lowering latency under compute-bound DLM inference.
Following the evaluation protocol of prior work~\cite{wu2025fastdllm}, we evaluate block-wise DLMs on representative benchmarks, including the GSM8K math benchmark~\cite{cobbe2021gsm8k} and the HumanEval coding benchmark~\cite{chen2021evaluating}.

As shown in Table~\ref{tab:diff-blocks-acc}, block-wise decoding consistently improves accuracy across both math and coding tasks compared to the naive DLM baseline (\texttt{None}).
While the block size $G$ is a hyperparameter whose optimal value may vary across tasks, all block-wise configurations outperform the naive DLM in our experiments.

Two factors likely contribute to this behavior.
First, block-wise decoding mitigates the excessive generation of end-of-text (\texttt{|EOT|}) tokens.
In naive DLMs, generation can collapse because the model can "cheat" by repeatedly filling \texttt{|EOT|} from later positions, since this is easy, leading to degenerate outputs.
Block-wise decoding helps alleviate this issue by enforcing parallel generation within the active block.
Second, block-wise decoding introduces a weak left-to-right inductive bias at the block level.
This ordering constraint can be beneficial for reasoning tasks, where a left-to-right structure naturally aligns with step-by-step reasoning processes.

\section{Implications of Reduction in Sampling Steps}
\label{appendix:reduce-steps}

\begin{figure}[t]
  \centering
  \includegraphics[width=\linewidth]{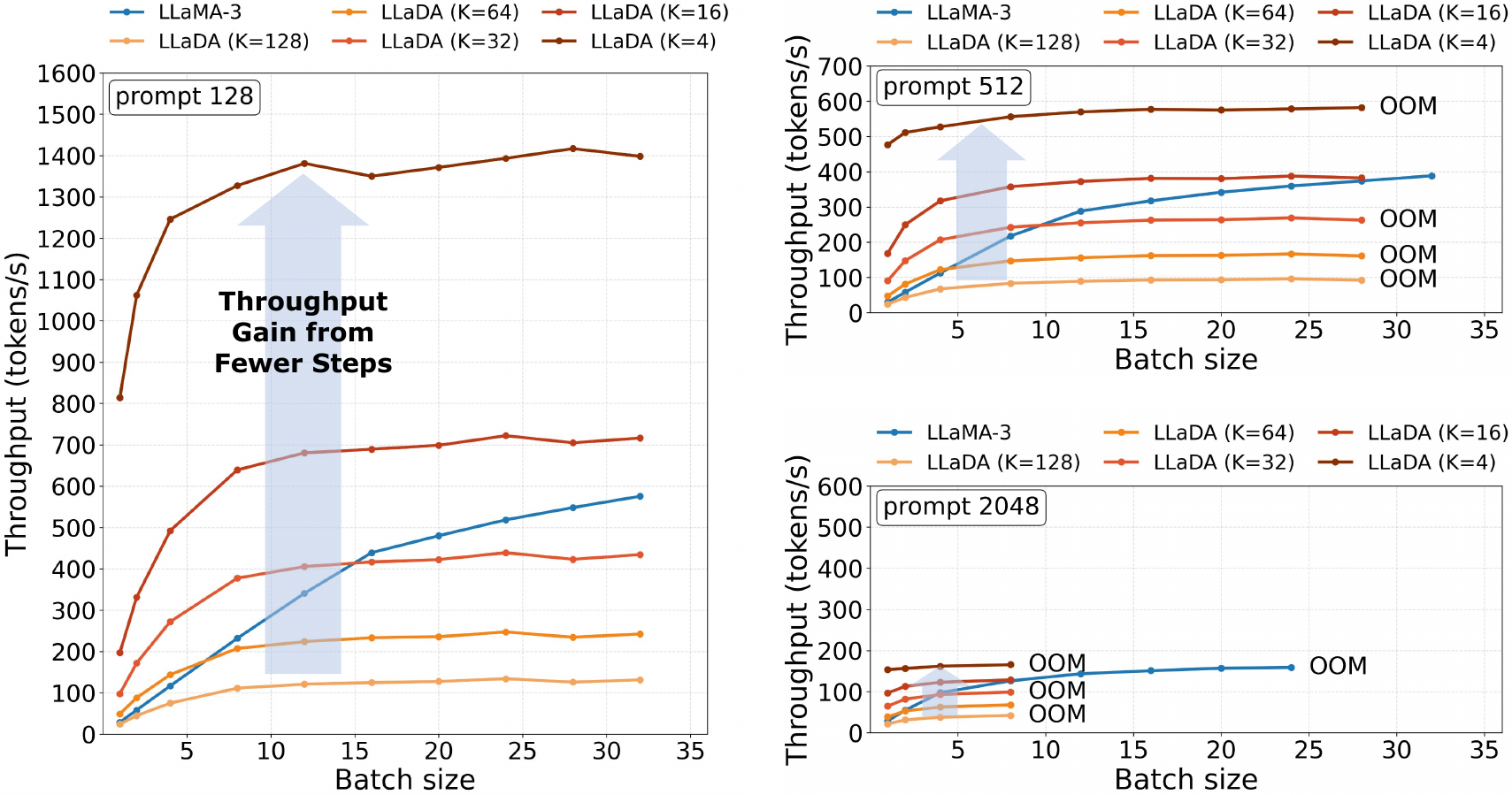}
  \vspace{-5mm}
  \caption{
    \textbf{Throughput vs. batch size for ARM and block-wise DLM with a varying number of sampling steps.} 
    The plots show the throughput of ARM and block-wise DLM for prompt length \(L_p\in\{128,512,2048\}\) with generation length $L_g=128$ and block size $G=32$. For LLaDA, $K$ denotes the number of sampling steps, which is varied from the baseline of 128 down to 4.
  }
  \label{fig:batched-runtime-analysis-diffsteps}
\end{figure}

As noted in our discussion, reducing the number of sampling steps $K$ is a critical path toward making DLMs competitive. To quantify this potential, we conduct a batch scalability experiment where we mechanically reduce the number of sampling steps $K$ from our default block-wise DLM configuration (block size $G=32$). 
Specifically, we reduce $K$ from its baseline of 128 (where $K=L_g$) down to 64, 32, 16, and 4. For example, when $K=4$, the generation length $L_g=128$ is divided into 4 blocks of 32 tokens each, and each block is denoised in a single refinement step.

\begin{table}[t]
  \centering
  \caption{
    Block-wise DLM performance on GSM8K and HumanEval benchmarks across different sampling steps $K$ (from 128 down to 4). 
    Generation length is fixed at $L_g = 128$ and block size $G=32$. 
    GSM8K is reported as exact-match (EM) accuracy, and HumanEval is reported as pass@1.
  }
  \label{tab:diff-steps-acc}
  \begin{tabular}{c|cc}
    \toprule
    \rowcolor{headergray}
    Sampling Steps $K$ & GSM8K (4-shot, EM) & HumanEval (0-shot, pass@1) \\
    \midrule
    128 & 73.6 & 35.4 \\
    64  & 70.6 & 32.3 \\
    32  & 61.7 & 17.7 \\
    16  & 29.7 & 9.1 \\
    4   & 6.1 & 3.7 \\
    \bottomrule
  \end{tabular}
\end{table}

In addition to throughput analysis, we also evaluate accuracy degradation when reducing steps without any further techniques. We again follow the evaluation setup of prior work~\cite{wu2025fastdllm} and test on representative benchmarks: the GSM8K math benchmark~\cite{cobbe2021gsm8k} and the HumanEval coding benchmark~\cite{chen2021evaluating}. As shown in Table~\ref{tab:diff-steps-acc}, naively reducing the number of steps results in severe accuracy drops, underscoring the necessity of advanced methods such as confidence-based multi-token finalization, autoregressive guidance, or distillation.

Figure~\ref{fig:batched-runtime-analysis-diffsteps} shows the results, with memory footprint remaining unchanged since only the number of iterations is reduced. For instance, with a 128-token prompt, we observe that although performance still plateaus around a batch size of 8, the peak throughput scales nearly proportionally as $K$ decreases.
These findings reinforce our earlier hypothesis in Section~\ref{sec:discussion}: in the \textit{small-batch} regime where DLMs can still benefit from batching, reducing the number of sampling steps (especially via methods that finalize multiple tokens per step) has the potential to enable DLMs to surpass ARMs in throughput, provided quality degradation is mitigated by advanced techniques.

\end{document}